\documentclass{llncs}

\usepackage[T1]{fontenc}
\usepackage{graphicx}
\usepackage{booktabs}
\usepackage{multirow}
\usepackage{microtype}
\usepackage{url}
\usepackage{array}
\usepackage[hidelinks]{hyperref}

\graphicspath{{figs/}}

\begin{document}

\title{Benchmarking Speaker Embedders in the Sub-Second Regime for Real-Time Voice AI Agents}
\titlerunning{Benchmarking Speaker Embedders in the Sub-Second Regime}

\author{Abhishek Gautam}
\authorrunning{A. Gautam}
\institute{\email{abhishek.gautam@iitg.ac.in}}

\maketitle

\begin{abstract}
Real-time Voice AI agents must separate the active user from intelligible background speech that survives conventional noise suppression. Background sources include television audio, family members, and colleagues. Production denoisers remove non-stationary environmental noise effectively. Background speech shares the spectral characteristics of the target and passes through, causing false acceptances in downstream speaker verification (SV) and automatic speech recognition (ASR). We benchmark open-source speaker embedders TitaNet-Large, TitaNet-Small, ECAPA-TDNN, CAM++, and SpeakerNet under a strict sub-second latency budget. Evaluation combines a controlled benchmark with a weighted human-annotated production set. Our controlled benchmark is built from real calls in which a second real speaker is injected into silence intervals at modulated amplitude. We report three findings. First, denoising before embedding is generally harmful and raises equal error rate (EER) across architectures. Second, truncating audio below the full voice-activity segment degrades every model. Third, roughly half of the TitaNet embedding is redundant for this task. Trailing dimensions encode population attributes such as language and gender. Such attributes carry little discriminative value in a demographically homogeneous user base. We distill these results into deployment guidance. Verification should run on raw audio with full voice-activity chunks. Thresholds should favor precision over F1. Truncated embeddings reduce the cost of similarity search.
\end{abstract}

\section{Introduction}\label{sec:introduction}
Conversational Voice AI systems operate in acoustically uncontrolled environments. Live calls routinely contain barge-in events and ambient chatter from televisions, colleagues, and family members. Voice agents must attribute each utterance to the target speaker or to a background speaker. Misattribution makes the agent answer questions the user never asked or treat a bystander remark as a user turn.

Production pipelines place a real-time denoiser such as DTLN~\cite{dtln} or DeepFilterNet~\cite{dfn} ahead of the voice activity detector. Denoisers excel at removing environmental noise such as fans and traffic. Background speech shares the spectral structure of target speech and frequently passes through intact. Conversation state becomes corrupted as a result. Speaker verification solves this by attributing each utterance to a target or background speaker.

Short utterances make this attribution acutely hard in live calls. Verification accuracy depends on sufficient speech to estimate a stable speaker embedding. Conversational turns are often brief backchannels such as yes, no, or okay that span only a few hundred milliseconds. Speaker embedders trained on multi-second utterances lose discriminative power sharply at these durations~\cite{poddar2018}. A naive remedy waits for more speech to accumulate before deciding. Waiting conflicts directly with the real-time budget of a live agent. Verification must therefore commit within a sub-second window where the embedding is least reliable. This tension between utterance length and latency defines the regime our study targets.

Our study benchmarks state-of-the-art speaker embedders for this exact regime of short utterances, telephony channels, and competing speech. Latency forms the first constraint. Pipelines must process audio chunks under one second to remain real-time. Accuracy forms the second constraint. Rejection of background speech must not reduce true acceptance of short target utterances.

\subsection{Related Work}
Prior work spans embedding architectures, speech enhancement, embedding interpretability, and overlap-aware speaker attribution. We review each in turn.

Speaker embedding research moved from i-vectors to deep frame-level d-vectors~\cite{dvector} and end-to-end training objectives~\cite{ge2e}. Snyder et al.~\cite{xvec} introduced x-vectors and established the TDNN plus statistics pooling template. ECAPA-TDNN~\cite{ecapa} added channel attention, multi-scale propagation, and attentive statistics pooling. TitaNet~\cite{titanet} uses depth-wise separable convolutions with squeeze-and-excitation and global context. TitaNet produces 192-D embeddings in a Large variant with 25.3\,M parameters and a Small variant with 6.4\,M. SpeakerNet~\cite{speakernet} is an earlier lightweight model from the same family. CAM++~\cite{campp} introduces context-aware masking on a densely connected TDNN backbone and offers strong accuracy at low inference cost. Such models are conventionally evaluated on VoxCeleb~\cite{voxceleb,voxceleb2} with multi-second utterances. Our setting of sub-second telephony chunks with competing speech is substantially harder. Embedding quality is known to degrade sharply as utterance duration shrinks~\cite{poddar2018}. Short duration explains absolute EERs far above the 1--2\% reported on VoxCeleb.

Speech enhancement advanced rapidly through the deep noise suppression challenge series~\cite{dns}. DTLN~\cite{dtln} is a dual-signal LSTM under 1\,M parameters operating in the STFT domain and a learned feature domain in real time. DeepFilterNet~\cite{dfn} performs deep filtering on full-band audio with ERB-scaled gains and a spectral refinement stage. Demucs~\cite{demucs} enhances directly in the waveform domain. Enhancement models optimize perceptual quality and intelligibility. Such objectives do not preserve the fine spectral structure of formants and harmonics on which verification depends. Our ablation in Section~\ref{subsec:denoising} quantifies this misalignment.

Interpretability studies show that speaker embeddings encode far more than identity. Session, channel, gender, and language information are linearly decodable~\cite{wang2017,raj2019}. Pluth et al.~\cite{sae} train sparse autoencoders on TitaNet embeddings and recover mono-semantic features for spoken language and hold music. Pluth et al.\ also steer these features independently of identity. Our dimensionality result in Section~\ref{subsec:dimensionality} provides complementary task-level evidence. Attribute directions contribute essentially nothing when the user population is homogeneous in those attributes.

Speaker attribution under overlapping speech is studied extensively in diarization~\cite{diarization}. Diarization systems decide offline over whole recordings. Our setting differs on both counts. We require an online decision within one second. We also exclude the overlapping-speech case and target only background speech that occupies its own VAD segment. Silero VAD~\cite{silero} produces these segments throughout this work. Each segment is then classified as target or background. Our benchmark replicates the exact production thresholds so that segmentation artifacts stay constant across all comparisons.

\subsection{Contributions}
Our contributions are fourfold. We construct a controlled benchmark of real calls with injected background speech. We ablate denoising front-ends across five verification architectures. We validate candidates on live production traffic including a chunk-duration sensitivity analysis. We show that the discriminative content of the 192-D TitaNet embedding concentrates in its first 92 dimensions. Section~\ref{sec:benchmark} details the benchmark design. Section~\ref{sec:experiments} reports experiments. Section~\ref{sec:discussion} distills deployment guidance and Section~\ref{sec:conclusion} concludes.

\section{Benchmark Design}\label{sec:benchmark}
This section describes how we build a controlled interference benchmark from real calls, configure voice activity detection, and derive denoised dataset variants.

\subsection{Controlled Interference Benchmark from Real Calls}
Our benchmark consists of 400 real customer calls and contains no synthetic speech. Each benchmark call combines two disjoint sources of genuine conversational audio. Base audio supplies the target speaker through a real call recording with the AI agent channel removed. Interference audio supplies the background speaker through speech from a different real speaker drawn from a separate pool of calls.

Enrollment uses the first 15 seconds of every base call. We leave this prefix untouched and compute the ground-truth target embedding from it. Production behaves the same way because the opening greeting is typically clean.

\begin{figure}[t]
\centering
\includegraphics[width=\textwidth]{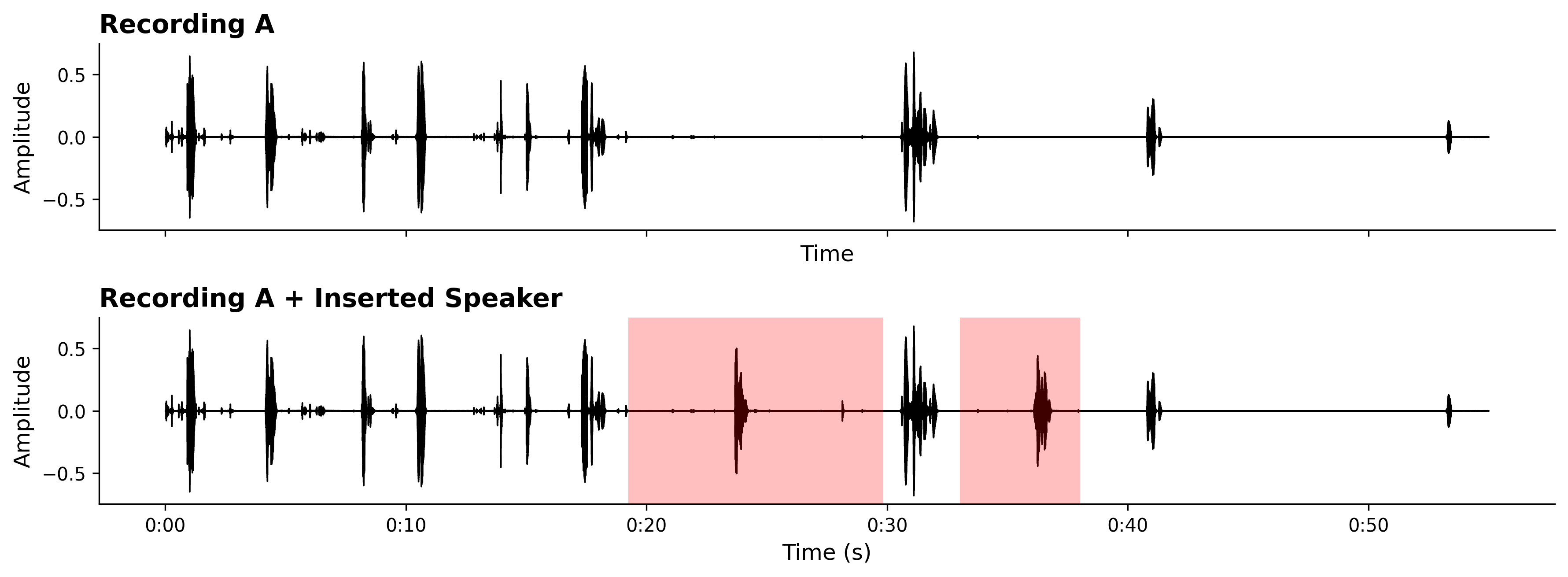}
\caption{Benchmark call construction with a clean 15\,s enrollment prefix and attenuated insertions of a second real speaker into silence intervals with 400\,ms safety buffers.}
\label{fig:construction}
\end{figure}

Injection begins beyond the 15-second mark. We scan the base audio for silence intervals and insert interference speech under two constraints. Each insertion sits inside a 400\,ms non-overlap safety buffer that prevents clipping of target speech. Amplitude modulation caps the inserted speech at 50\% of the base signal's maximum amplitude and simulates acoustic distance. Our corpus therefore contains real conversations with precisely controlled and realistically attenuated background interference. Figure~\ref{fig:construction} illustrates the construction.

\subsection{Voice Activity Detection Configuration}
Silero VAD~\cite{silero} segments all audio with the exact production parameters. Threshold equals 0.55. Minimum silence duration equals 300\,ms. Speech padding equals 30\,ms. Matching parameters ensure that benchmark segmentation mirrors live behavior.

\subsection{Dataset Variants and Denoiser Side-Effects}
Denoising affects segmentation as well as spectral content. We processed the corpus through two candidate denoisers and obtained the three dataset variants in Table~\ref{tab:variants}. DeepFilterNet suppresses roughly 360 soft utterances below the VAD threshold. DTLN introduces spectral artifacts that fragment long utterances into more and shorter chunks. Studies that combine enhancement with verification rarely report this confound. Section~\ref{sec:experiments} evaluates five embedders on these variants and on live traffic.

\begin{table}[t]
\centering
\caption{Dataset variants after VAD segmentation.}
\label{tab:variants}
\footnotesize
\begin{tabular}{@{}lr>{\raggedright\arraybackslash}p{6.2cm}@{}}
\toprule
\textbf{Variant} & \textbf{Utts.} & \textbf{Observation} \\
\midrule
Original (raw) & 7{,}607 & Baseline and ground truth for natural background-speech detection. \\
DeepFilterNet & 7{,}247 & VAD missed roughly 360 background utterances because soft background speech fell below the VAD threshold. \\
DTLN & 7{,}971 & Highest count because spectral artifacts fragment long utterances into multiple shorter chunks. \\
\bottomrule
\end{tabular}
\end{table}

\section{Experiments and Results}\label{sec:experiments}
Section~\ref{sec:experiments} presents results in five parts. The first two subsections cover the controlled benchmark. The next two cover live production validation. The final subsection covers the embedding-dimensionality ablation. Equal Error Rate (EER) serves as the primary metric for target and background separation. Production results additionally report estimated false positives and false negatives at the EER operating point.

\subsection{Baseline Performance on Raw Audio}
\begin{figure}[t]
\centering
\includegraphics[width=\textwidth]{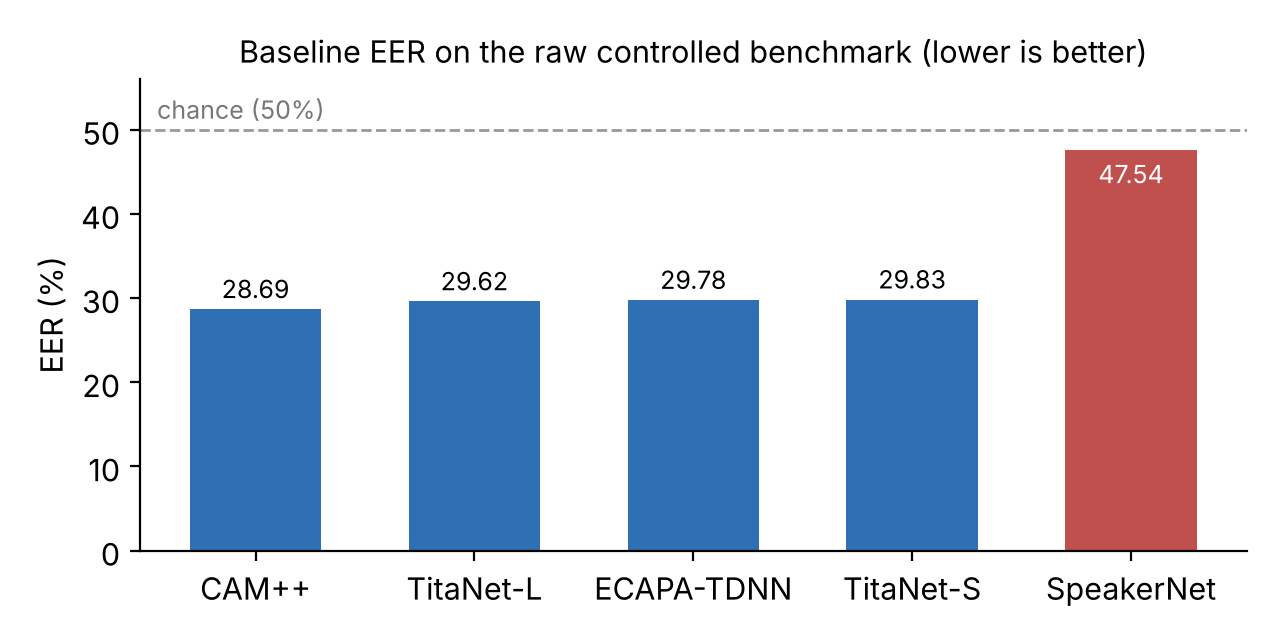}
\caption{Baseline EER on the raw controlled benchmark. CAM++ leads. TitaNet variants and ECAPA-TDNN sit within 0.21 points of one another, about one point behind CAM++. SpeakerNet is near chance.}
\label{fig:baseline}
\end{figure}

Baseline experiments use raw non-denoised audio with short inputs under one second. Table~\ref{tab:baseline} and Figure~\ref{fig:baseline} summarize the results. CAM++ generalizes best to raw short utterances with 28.69\% EER. TitaNet-Large, TitaNet-Small, and ECAPA-TDNN perform within 0.21 points of one another and roughly one point behind CAM++. Scaling model size therefore yields diminishing returns for this task. SpeakerNet sits at 47.54\% EER, performs near chance, and is unsuitable for separating background speech in short segments.

\begin{table}[t]
\centering
\caption{Baseline results on raw non-denoised audio with short inputs under 1\,s.}
\label{tab:baseline}
\footnotesize
\begin{tabular}{@{}lcc@{}}
\toprule
\textbf{Model} & \textbf{EER (\%)} & \textbf{Best F1} \\
\midrule
\textbf{CAM++} & \textbf{28.69} & \textbf{0.8315} \\
TitaNet-Large & 29.62 & 0.8076 \\
ECAPA-TDNN & 29.78 & 0.8200 \\
TitaNet-Small & 29.83 & 0.8273 \\
SpeakerNet & 47.54 & 0.8268 \\
\bottomrule
\end{tabular}
\end{table}

\subsection{Effect of Denoising Front-Ends}\label{subsec:denoising}
Our working hypothesis held that denoising would remove background artifacts and improve verification. Results largely contradict it. Table~\ref{tab:denoise} lists EER for every embedder under each front-end. DeepFilterNet degraded EER universally, by between 2.03 and 4.89 points for the four competitive models. CAM++ moved from 28.69\% to 30.72\% and TitaNet-Small moved from 29.83\% to 34.72\%. DeepFilterNet suppresses noise partly by smoothing spectral fine-structure. Formant detail is precisely what verification models discriminate on. Enhancement therefore homogenizes voice characteristics and makes target and background speakers harder to separate.

\begin{figure}[!t]
\centering
\includegraphics[width=0.78\textwidth]{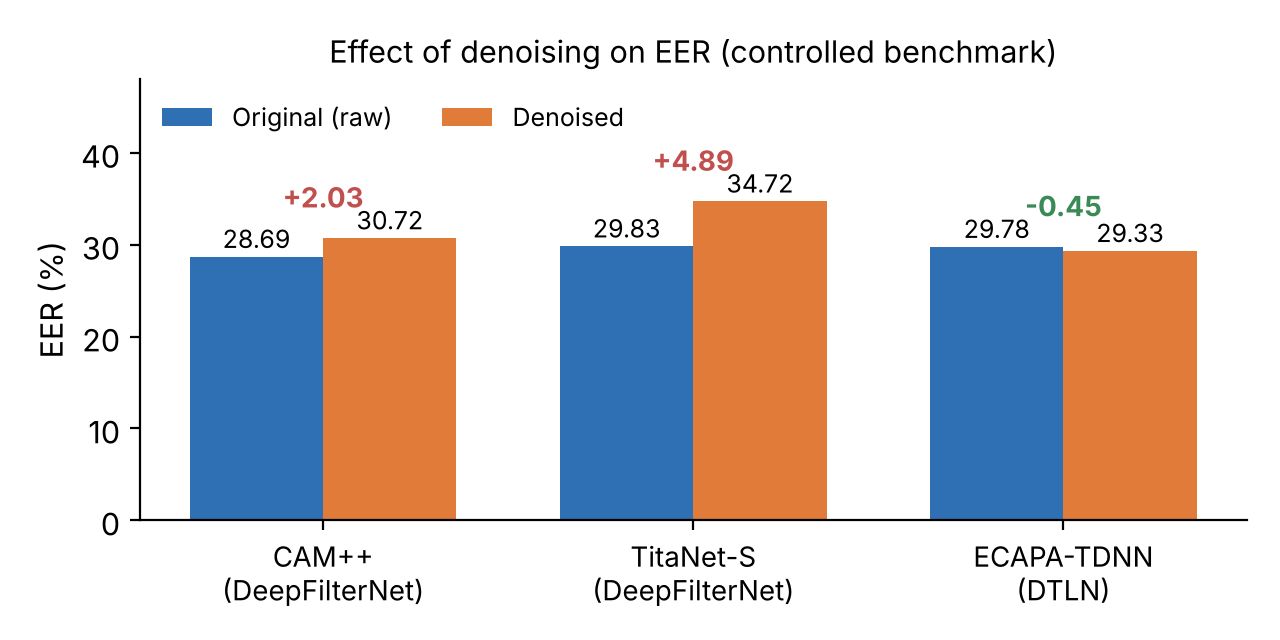}
\caption{Denoising ablation on the controlled benchmark. Deltas above each pair mark the change in EER after denoising. ECAPA-TDNN with DTLN is the single improving configuration and still does not match raw-audio CAM++.}
\label{fig:denoise}
\end{figure}

DTLN shows an architecture-dependent effect. ECAPA-TDNN was the only competitive model to improve under denoising and moved from 29.78\% to 29.33\%. SpeakerNet also shifted marginally, by 0.13 points, but remains near chance. ECAPA-TDNN appears more robust to spectral-masking artifacts than its peers. Critically the best denoised configuration of ECAPA-TDNN with DTLN remains worse than raw-audio CAM++ at 28.69\%. Denoising never enters the Pareto frontier as Figure~\ref{fig:denoise} shows. Verification should therefore consume raw audio. Live traffic in the next part tests whether benchmark rankings transfer to production.

\begin{table}[t]
\centering
\caption{EER (\%) on the controlled benchmark for each front-end. Denoised variants also change the number of VAD segments (Table~\ref{tab:variants}).}
\label{tab:denoise}
\footnotesize
\begin{tabular}{@{}lccc@{}}
\toprule
\textbf{Model} & \textbf{Raw} & \textbf{DeepFilterNet} & \textbf{DTLN} \\
\midrule
CAM++ & \textbf{28.69} & 30.72 & 30.38 \\
TitaNet-Large & 29.62 & 33.18 & 29.96 \\
ECAPA-TDNN & 29.78 & 32.16 & 29.33 \\
TitaNet-Small & 29.83 & 34.72 & 31.34 \\
SpeakerNet & 47.54 & 51.81 & 47.41 \\
\bottomrule
\end{tabular}
\end{table}

\subsection{Live Production Validation}
\begin{figure}[t]
\centering
\includegraphics[width=\textwidth]{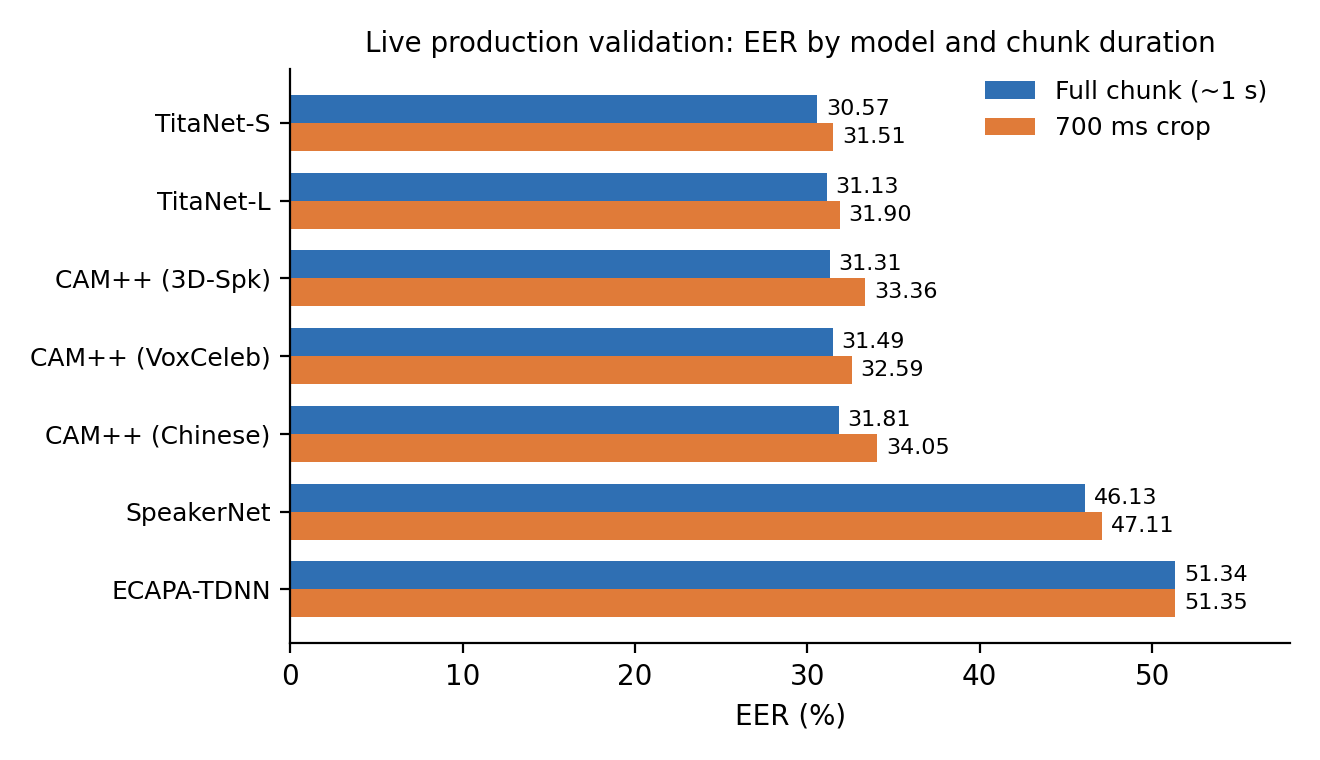}
\caption{Production EER by model and chunk duration. TitaNet-Small with full chunks takes the top spot. ECAPA-TDNN collapses to chance on live data despite acceptable benchmark performance.}
\label{fig:prod}
\end{figure}

Live validation evaluates candidate models on real traffic with channel noise, codec artifacts, and heterogeneous microphones. Human annotators labeled utterance pairs as same speaker, different speaker, or noise. Statistical weighting reconstructs the full production distribution because only a subset of traffic is labeled. Weighted samples total \textbf{11{,}070} with 6{,}661 target at 60.2\%, 1{,}328 background at 12.0\%, and 1{,}200 noise at 10.8\%. Table~\ref{tab:prod} and Figure~\ref{fig:prod} report the results.

\begin{table}[t]
\centering
\caption{Live production validation ranked by EER on the weighted distribution. Full-duration chunks dominate 700\,ms crops for every architecture.}
\label{tab:prod}
\footnotesize
\setlength{\tabcolsep}{2pt}
\resizebox{\textwidth}{!}{%
\begin{tabular}{@{}llccccccc@{}}
\toprule
\textbf{Model} & \textbf{Duration} & \textbf{EER (\%)} & \textbf{Thresh @ EER} & \textbf{F1} & \textbf{Recall} & \textbf{FP (est.)} & \textbf{FN (est.)} & \textbf{Acc.} \\
\midrule
\textbf{TitaNet-Small} & \textbf{Full} & \textbf{30.57} & \textbf{0.3336} & \textbf{0.7325} & \textbf{0.6953} & \textbf{1{,}352} & \textbf{2{,}029} & \textbf{0.6945} \\
TitaNet-Large & Full & 31.13 & 0.3027 & 0.7269 & 0.6885 & 1{,}371 & 2{,}075 & 0.6887 \\
CAM++ (3D-Speaker) & Full & 31.31 & 0.3256 & 0.7254 & 0.6872 & 1{,}382 & 2{,}083 & 0.6869 \\
CAM++ (VoxCeleb) & Full & 31.49 & 0.2751 & 0.7237 & 0.6854 & 1{,}389 & 2{,}095 & 0.6852 \\
TitaNet-Small & 700\,ms & 31.51 & 0.3108 & 0.7236 & 0.6853 & 1{,}391 & 2{,}096 & 0.6850 \\
CAM++ (Chinese) & Full & 31.81 & 0.4045 & 0.7205 & 0.6815 & 1{,}401 & 2{,}121 & 0.6818 \\
TitaNet-Large & 700\,ms & 31.90 & 0.2727 & 0.7200 & 0.6814 & 1{,}408 & 2{,}122 & 0.6811 \\
CAM++ (VoxCeleb) & 700\,ms & 32.59 & 0.2496 & 0.7129 & 0.6729 & 1{,}431 & 2{,}178 & 0.6739 \\
CAM++ (3D-Speaker) & 700\,ms & 33.36 & 0.2962 & 0.7061 & 0.6663 & 1{,}470 & 2{,}223 & 0.6663 \\
CAM++ (Chinese) & 700\,ms & 34.05 & 0.3746 & 0.7000 & 0.6600 & 1{,}503 & 2{,}265 & 0.6596 \\
SpeakerNet & Full & 46.13 & 0.8040 & 0.5844 & 0.5389 & 2{,}035 & 3{,}071 & 0.5387 \\
SpeakerNet & 700\,ms & 47.11 & 0.7931 & 0.5744 & 0.5284 & 2{,}075 & 3{,}141 & 0.5288 \\
ECAPA-TDNN & Full & 51.34 & 0.0012 & 0.5332 & 0.4871 & 2{,}265 & 3{,}416 & 0.4867 \\
ECAPA-TDNN & 700\,ms & 51.35 & 0.0019 & 0.5332 & 0.4873 & 2{,}267 & 3{,}415 & 0.4867 \\
\bottomrule
\end{tabular}%
}
\end{table}

Three observations stand out. Rankings invert relative to the controlled benchmark. TitaNet-Small on full chunks achieves the lowest EER at 30.57\% with the highest F1 at 0.7325 and overtakes both TitaNet-Large and the benchmark winner CAM++. Smaller models can generalize better to real channel and codec conditions than benchmark-tuned larger ones. ECAPA-TDNN collapses to near-chance performance at 51.3\% EER with a near-zero operating threshold despite mid-pack benchmark results. Controlled benchmarks alone are therefore insufficient for model selection. CAM++ pretrained on 3D-Speaker~\cite{3dspeaker} edges out the VoxCeleb and Chinese variants yet all CAM++ versions trail TitaNet on this distribution.

\subsection{Chunk-Duration Sensitivity}
\begin{figure}[t]
\centering
\includegraphics[width=0.96\textwidth]{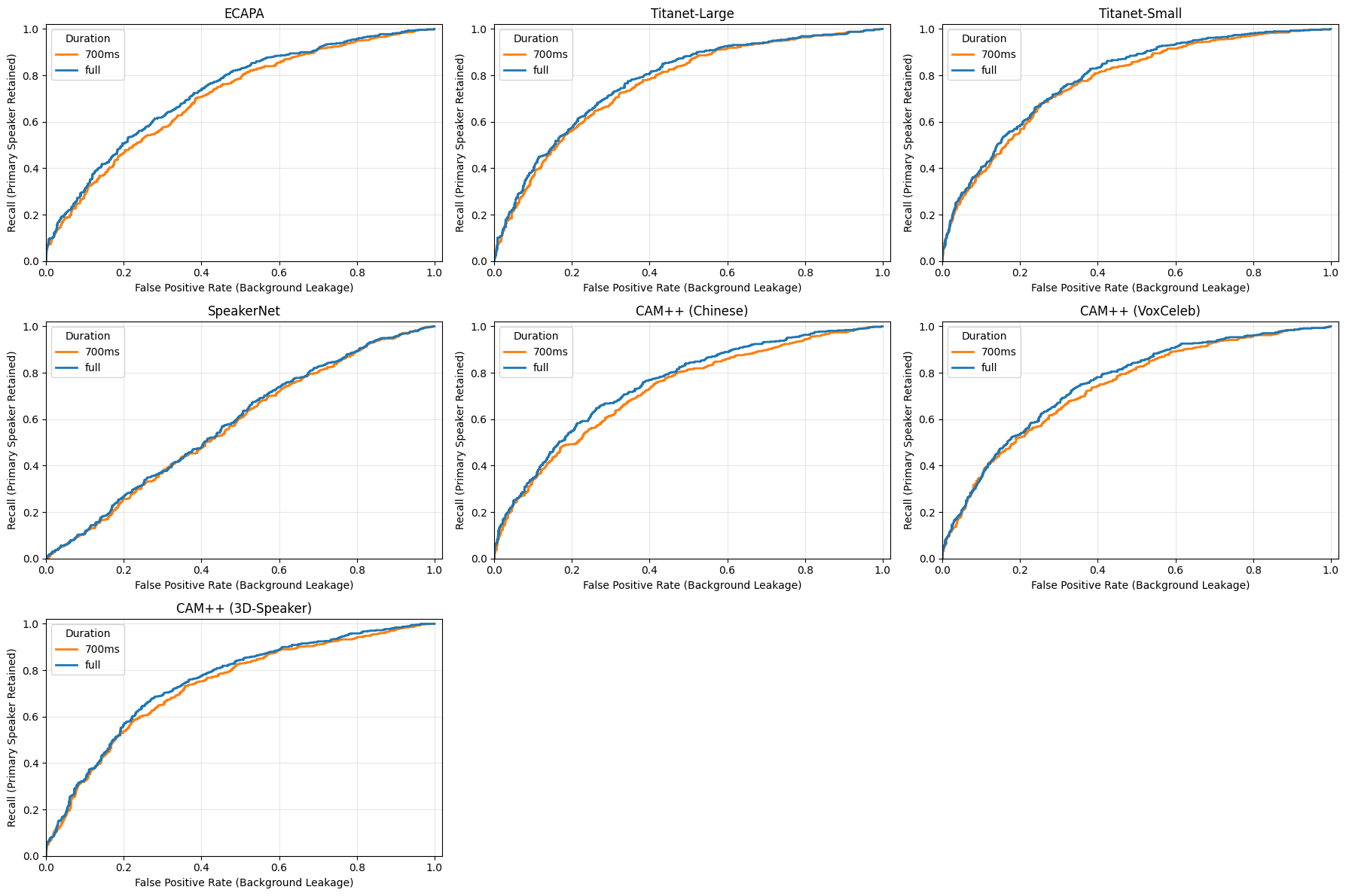}
\caption{ROC curves for full chunks in blue and 700\,ms crops in orange. Full-duration curves dominate particularly at low false-positive rates.}
\label{fig:roc}
\end{figure}

Duration experiments ask whether audio can be truncated to a fixed 700\,ms without accuracy loss. Results indicate it cannot. Full-duration chunks up to roughly one second consistently outperform 700\,ms crops. TitaNet-Small degrades by about 1.0 point. TitaNet-Large degrades by about 0.8. CAM++ shows the highest duration sensitivity and degrades by about 2.0 with 3D-Speaker weights and 2.2 with Chinese weights. ROC curves in Figure~\ref{fig:roc} confirm that the full-duration curve envelops the 700\,ms curve especially in the operationally critical low-FPR region. Additional context of roughly 200 to 300\,ms in the full VAD chunk stabilizes the embedding. Stability matters disproportionately for the lightweight TitaNet-Small. Full variable-length VAD segments should therefore be retained.

\subsection{Embedding Dimensionality Ablation}\label{subsec:dimensionality}
\begin{figure}[t]
\centering
\includegraphics[width=\textwidth]{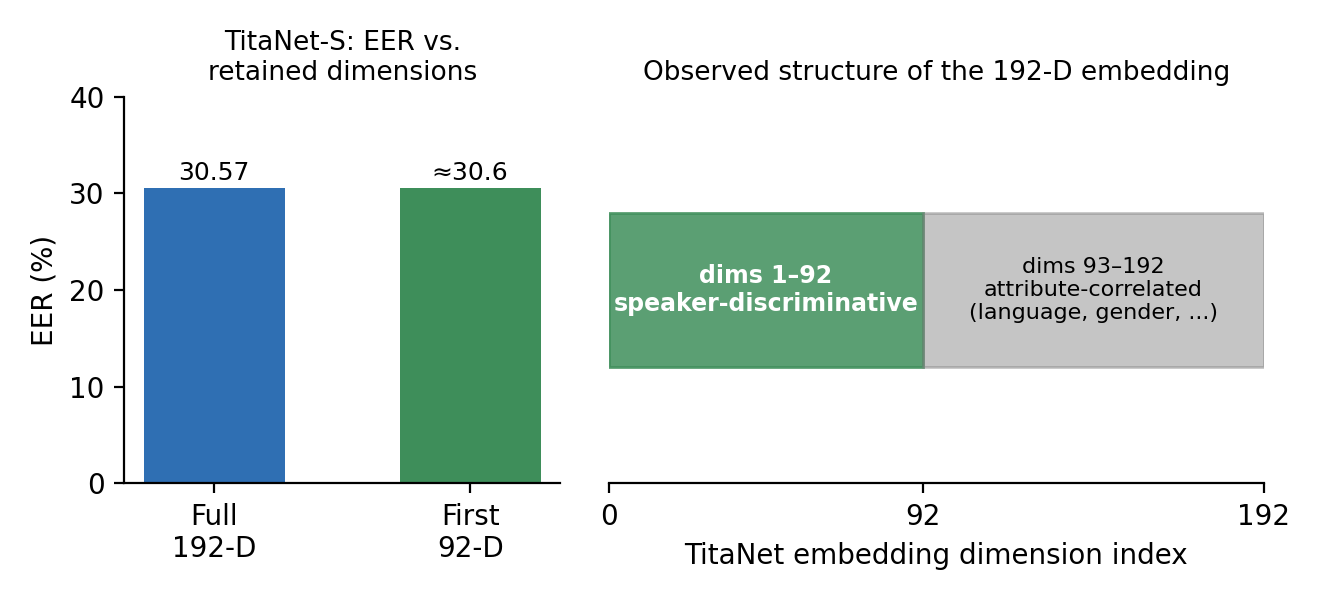}
\caption{Truncating TitaNet-Small embeddings to the first 92 dimensions leaves EER essentially unchanged (left). Leading dimensions carry speaker-discriminative information while trailing dimensions correlate with population attributes such as language and gender (right).}
\label{fig:dims}
\end{figure}

Dimensionality experiments ask how discriminative information distributes across the 192 dimensions of the TitaNet embedding. Interpretability results on TitaNet~\cite{sae} motivated this question. We re-ran verification with only the \textbf{first 92 dimensions} on both enrollment and test sides and computed cosine similarity in the truncated space. Retaining dimensions 1 to 92 did not degrade EER relative to the full embedding. Measured EER matched the full-dimensional figure within noise. Task-relevant information therefore concentrates in the leading half of the vector as Figure~\ref{fig:dims} shows.

Qualitative inspection indicates that trailing dimensions are dominated by coarse population attributes such as spoken language and gender. Probing studies report that language and gender are linearly decodable from speaker embeddings~\cite{wang2017,raj2019}. Pluth et al.~\cite{sae} isolate mono-semantic language and music features inside TitaNet embeddings and steer them independently of identity. Our user population is heavily skewed toward male Hindi speakers. Attribute-correlated directions therefore provide almost no separating power between target and background speakers who typically share both attributes. Truncation discards them without penalty.

Caution is warranted because this is an observational ablation rather than a causal attribution. We have shown that trailing dimensions are dispensable for this distribution. We have not pinned individual dimensions to individual features. SAE methodology~\cite{sae} or per-dimension probing~\cite{raj2019} could map dimensions to attributes explicitly. Transfer to demographically diverse populations should be verified and will likely fail where language and gender do discriminate. Practically a 92-D representation halves similarity computation and storage cost. Savings compound at production scale. Section~\ref{sec:discussion} consolidates these findings into deployment guidance.

\section{Discussion}\label{sec:discussion}
Section~\ref{sec:discussion} converts the experimental threads into a deployment recipe and states broader lessons and limitations.

Front-end choice forms the first decision. Speaker verification should consume the raw non-denoised stream. DeepFilterNet raised EER for every top-tier embedder, by between 2.03 points for CAM++ and 4.89 for TitaNet-Small. DTLN with ECAPA-TDNN produced the only improvement at 0.45 points and still trails the best raw configuration by 0.64 points. Pipelines that need enhancement for transcription should branch after capture. Denoised audio feeds ASR while verification consumes the unprocessed signal.

Model and chunk duration form the second decision and live data should outrank benchmark data here. TitaNet-Small on full VAD chunks is our recommended configuration. TitaNet-Small attains the lowest production EER at 30.57\% with the highest F1 at 0.7325 while being roughly four times smaller than TitaNet-Large. CAM++ led the controlled benchmark by more than a full point yet slipped behind both TitaNet variants on live traffic. ECAPA-TDNN degraded to near chance under real channel and codec conditions. Chunk duration follows the same logic. Fixing the input at 700\,ms costs between 0.8 and 2.2 EER points for the competitive architectures. Losses concentrate in the low false-positive region where barge-in suppression actually operates. Full variable-length VAD segments justify their modest latency premium.

Operating threshold and representation form the third decision. Accepting a bystander utterance corrupts conversation state. Rejecting a genuine user utterance merely triggers a cheap clarification. Error costs are therefore asymmetric and the cosine threshold should favor precision. Thresholds should sit at or above the production EER operating point of roughly 0.33 for TitaNet-Small. Table~\ref{tab:prod} still estimates roughly 1{,}350 non-target acceptances at that point on the weighted distribution. Thresholds tuned for F1 sit far lower on the similarity scale and inflate false acceptances substantially. Dimensionality results are directly actionable as well. Truncating TitaNet embeddings to the first 92 dimensions halves every similarity computation and every stored enrollment vector with no measured EER penalty on this population.

Two broader lessons generalize beyond this system. Perceptual-quality objectives and discriminability objectives diverge. Components optimized to make audio sound clean can make speakers measurably less separable. Controlled benchmarks are necessary but not sufficient for model selection. ECAPA-TDNN collapsed from 29.8\% EER on the benchmark to 51.3\% in production. Channel and codec effects can invert a model ranking entirely. Candidates must be validated on traffic from the target deployment before adoption.

\subsection{Limitations and Future Work}
Several caveats bound these conclusions. Absolute EERs near 30\% reflect the difficulty of verifying a speaker from a single sub-second telephony utterance. Numbers are not comparable to multi-second VoxCeleb protocols. Score aggregation across consecutive chunks of the same call is the natural mitigation and remains future work. Overlapping speech lies outside our segment-level scope. We classify each VAD segment as target or background and never measure concurrent speakers inside one segment. Our injection protocol caps interference at 50\% amplitude inside silence intervals. Fully overlapped cocktail-party speech and louder interferers are left to future work. Dimensionality findings are specific to a male-skewed Hindi-speaking population. Mapping from trailing dimensions to attributes is presently observational. SAE-based attribution~\cite{sae} or per-dimension probing~\cite{raj2019} would make the mapping causal. Transfer to demographically diverse populations should not be assumed. In-domain fine-tuning of TitaNet on telephony data reached sub-1\% EER in the telephony domain in~\cite{sae} and is the most direct lever for closing the remaining accuracy gap.

\section{Conclusion}\label{sec:conclusion}
We presented a benchmarking study of state-of-the-art speaker embedders for low-latency Voice AI under background-speech interference. Denoising front-ends remain valuable for ASR yet are net harmful to verification. Raw-audio CAM++ dominated all denoised configurations on the controlled benchmark. TitaNet-Small on full VAD chunks emerged as the production winner. Chunk truncation to 700\,ms degraded every architecture. F1-optimal thresholds proved operationally unsafe due to false-positive volume. Discriminative content of the 192-D TitaNet embedding concentrates in its first 92 dimensions for our deployment population. Trailing dimensions encode language and gender attributes. Our observation echoes recent interpretability work on voice embeddings and yields an immediate twofold efficiency gain.

% Optional for arXiv. Uncomment and fill in if you have funding to acknowledge
% or an employer/data-provider relationship to disclose.
% \begin{credits}
% \subsubsection{\ackname} ...
% \subsubsection{\discintname} ...
% \end{credits}

\end{document}